\documentclass[10pt,twocolumn,letterpaper]{article}
\usepackage{cvpr}              
\usepackage{epsfig}
\usepackage{amsmath}
\usepackage{amssymb}
\usepackage{graphicx}
\usepackage{tikz}
\usepackage{comment}
\usepackage{color}
\usepackage{pifont}
\usepackage{epsfig}
\usepackage{graphics}
\usepackage{multirow}
\usepackage{algorithm}
\usepackage{algorithmicx}
\usepackage{subcaption}
\usepackage{sidecap}
\usepackage{wrapfig}
\usepackage{enumitem}
\usepackage{colortbl}
\usepackage{arydshln,xcolor,array}
\usepackage{multirow}
\usepackage{esvect}
\usepackage{bbm}
\usepackage{graphicx}
\usepackage{booktabs}
\newcommand*\colourcheck[1]{%
  \expandafter\newcommand\csname #1check\endcsname{\textcolor{#1}{\ding{51}}}%
}
\newcommand*\colourcross[1]{%
  \expandafter\newcommand\csname #1cross\endcsname{\textcolor{#1}{\ding{55}}}%
}
\colourcheck{green}
\colourcross{red}
\let\origtau\tau 
\renewcommand{\tau}{\scalebox{1.2}{$\origtau$}}

\definecolor{cvprblue}{rgb}{0.21,0.49,0.74}
\usepackage[pagebackref,breaklinks,colorlinks,citecolor=cvprblue]{hyperref}

\def\paperID{142} 
\def\confName{3DV\xspace}
\def\confYear{2025\xspace}
\newcommand{\model}[0]{SCENIC~}
\newcommand{\vect}[1]{\boldsymbol{#1}}
\newcommand{\mat}[1]{\mathbf{#1}}

\newcommand{\removespace}{\hspace{-\fontdimen2\font plus -\fontdimen3\font minus -\fontdimen4\font}}

\newcommand{\human}{\mat{H}}
\newcommand{\clip}{\mat{T}}
\newcommand{\goal}{\mat{G}}
\newcommand{\scene}{\mat{S}}
\newcommand{\object}{\mat{O}}
\newcommand{\condition}{\mat{C}}

\newcommand{\poseposition}{\vect{J}_{p}}

\newcommand{\poserotation}{\vect{J}_{r}}
\newcommand{\contact}{\vect{c}}

\newcommand{\goalpos}{\vect{g}_{p}}
\newcommand{\goaldir}{\vect{g}_{r}}

\usepackage{graphicx} 
\usepackage{caption} 
\usepackage[utf8]{inputenc} 

\title{SCENIC: Scene-aware Semantic Navigation with Instruction-guided Control}
\begin{document}

\author{
    Xiaohan Zhang$^{1,2}$, Sebastian Starke$^{3}$, 
    Vladimir Guzov$^{1,2}$, \\ Zhensong Zhang$^{4}$, Eduardo P\'erez-Pellitero$^{4}$, Gerard Pons-Moll$^{1,2}$ \\
}

\makeatletter
\let\@oldmaketitle\@maketitle 
\renewcommand{\@maketitle}{%
    \@oldmaketitle 
    \vspace{-1.5em} 
    \begin{center}
        \normalsize
        $^1$Tübingen AI Center, University of Tübingen \\
        $^2$Max Planck Institute for Informatics, Saarland Informatics Campus \\
        $^3$Meta Reality Labs Research\\
        $^4$Huawei Noah's Ark Lab \\
        \vspace{2em} 
        \includegraphics[width=.9\textwidth]{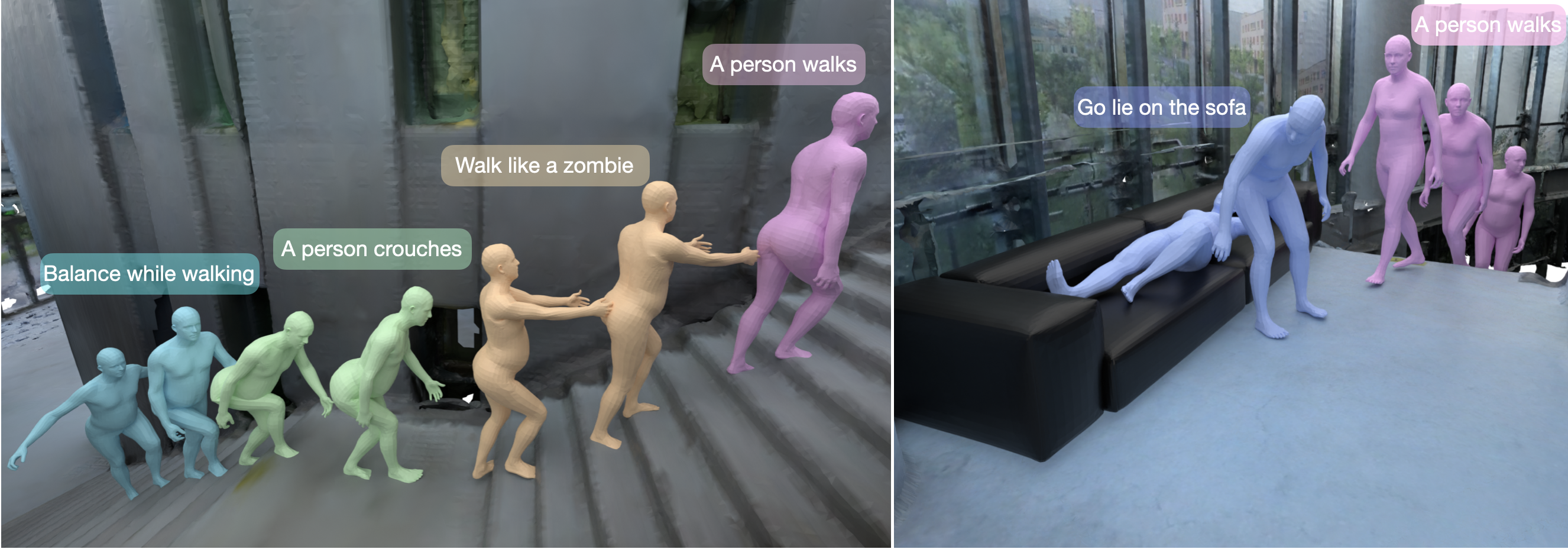} 
    \end{center}
    \refstepcounter{figure}
    Figure 1: ~\model is a text-conditioned scene interaction model. It adapts to complex scenes with varying terrains and also supports user-specified semantic control with natural language. Given a 3D scene, our model takes as cues of user-specified trajectory as sub-goals, and text.~\textit{We encourage the readers to watch the supplementary video.}
    \label{fig:teaser}
    \bigskip
}
\makeatother

\maketitle

\begin{abstract}

Synthesizing natural human motion that adapts to complex environments while allowing creative control remains a fundamental challenge in motion synthesis. Existing models often fall short, either by assuming flat terrain or lacking the ability to control motion semantics through text. To address these limitations, we introduce~\model\removespace, a diffusion model designed to generate human motion that adapts to dynamic terrains within virtual scenes while enabling semantic control through natural language. The key technical challenge lies in simultaneously reasoning about complex scene geometry while maintaining text control. This requires understanding both high-level navigation goals and fine-grained environmental constraints. The model must ensure physical plausibility and precise navigation across varied terrain, while also preserving user-specified text control, such as ``carefully stepping over obstacles" or ``walking upstairs like a zombie." Our solution introduces a hierarchical scene reasoning approach. At its core is a novel scene-dependent, goal-centric canonicalization that handles high-level goal constraint, and is complemented by an ego-centric distance field that captures local geometric details. This dual representation enables our model to generate physically plausible motion across diverse 3D scenes. By implementing frame-wise text alignment, our system achieves seamless transitions between different motion styles while maintaining scene constraints. Experiments demonstrate our novel diffusion model generates arbitrarily long human motions that both adapt to complex scenes with varying terrain surfaces and respond to textual prompts. Additionally, we show~\model can generalize to four real-scene datasets. Our code, dataset, and models will be released at \url{https://virtualhumans.mpi-inf.mpg.de/scenic/}.

\vspace{-10pt}
\end{abstract} 

\section{Introduction}

Humans navigate complex environments effortlessly, adapting to varied terrains while performing diverse motions. This fundamental ability to synthesize natural human motion in complex environments~\cite{trumans,jiang2024autonomous,mir23origin,Zhao:ICCV:2023} is crucial for numerous applications ranging from gaming to embodied AI. For instance, how can we make virtual characters seamlessly ``step over obstacles before sitting" or ``walk upstairs like a zombie" (Figure~\ref{fig:teaser}). Fundamentally, this requires both scene understanding and semantic control. While recent works have made progress in either text-controlled human motion synthesis~\cite{tevet2023human,shafir2023human,xie2024omnicontrol} or motion adaptation to simplified environments~\cite{yi2024tesmo,li2024chois}, they struggle with complex scenarios. Even methods that can adapt to uneven terrain~\cite{pfnn,rempeluo2023tracepace,luo2024universal} lack flexible semantic control through natural language. This work bridges this gap by introducing a unified diffusion-based framework that simultaneously handles complex scene geometry and text-based semantic control.

Synthesizing scene-aware semantic motion faces three fundamental challenges. First, the model must generate motion that precisely adapts to complex environment constraints, avoiding penetration, while maintaining natural contact with uneven surfaces, and reaching specific targets. Furthermore, unlike previous approaches that handle either scene geometry or semantic control in isolation, combining both requires sophisticated reasoning about how different motion styles interact with varied terrain features. Last, traditional approaches require extensive paired motion-scene data, which is expensive to acquire due to tracking difficulties and does not scale well to diverse environments.

Our key insight is that complex scene-aware motion synthesis can be decomposed into hierarchical reasoning levels, similar to how humans approach navigation tasks. At the high level, we synthesize motion in a goal-centric canonical coordinate frame, enabling the model to learn target-reaching behaviors naturally. At a more granular level, we take inspiration from recent 3D generation work~\cite{Chan2022eg3d} that encodes 3D spatial features with 2D planar encoding. We represent detailed scene geometry through a human-centered distance field representation~\cite{rempeluo2023tracepace,luo2024universal}. This efficient representation enables comprehensive reasoning about local scene features, including terrain variations and obstacles. To provide semantic control, we align text and motion on a frame-wise basis, allowing for dynamic instruction changes while ensuring smooth transitions. To address data efficiency, we exploit the compositional nature of human motion, training on short motion segments~\cite{mir23origin,trumans,jiang2024autonomous} that can be efficiently augmented by automatically fitting varied terrain surfaces.

With these solutions, we propose the first model which is scene-aware and can be controlled with fine-grained natural language. Experiments demonstrate~\model handles complex scene geometry through precise scene-aware adaptation across four real-scene datasets including Replica~\cite{replica19arxiv}, Matterport3D~\cite{Matterport3D}, HPS~\cite{HPS}, and LaserHuman~\cite{cong2024laserhuman}. Moreover,~\model supports seamless transition between ten distinct motion semantics including ``crouching", ``climbing", ``hopping", ``jumping'', and ``balancing", and can adapt to complicated instruction such as ``walking upstairs like a zombie''. Empirically, our model achieves the best in terms of satisfying the scene and goal constraints, and motion quality. Qualitatively, our model is preferred by \emph{75.6\%} of participants over state-of-the-art alternatives (more details see Table~\ref{tab:my_label}).

The key contributions of our work include:
\begin{enumerate}
    \item We introduce the first unified method for 3D scene-aware human motion synthesis, capable of handling complex terrains like stairs, steps, or slopes, while also enabling fine-grained control through textual prompts.
    \item Our novel diffusion model leverages hierarchical scene reasoning, efficiently handles complex 3D environments while maintaining plausibility. Its effectiveness is validated across four diverse real-world datasets.
    \item A scalable approach to synthesizing continuous human navigation in 3D scenes, which can be integrated with an object-interaction model, as shown in Figure~\ref{fig:teaser}.
\end{enumerate}

\section{Related Work}
\subsection{Text-guided Motion Diffusion.}

Recent years have seen remarkable progress in human motion synthesis, driven by the emergence of diffusion models~\cite{tevet2023human, shafir2023human,zhang2022motiondiffuse,zhang2023remodiffuse,chen2023executing,dabral2022mofusion,HoangGGM24,humantomato,ma2024richcat,zhang2024tedi,zhou2023emdm,KongGLMW23} and comprehensive motion capture datasets like AMASS~\cite{amass}. The integration of action labels and language descriptions through datasets such as BABEL~\cite{BABEL:CVPR:2021} and HumanML3D~\cite{Guo_2022_CVPR} has enabled increasingly sophisticated control over generated motions. Recent work has explored various aspects of motion synthesis, including two-person interactions~\cite{ghosh2024remos,liang2024intergen,tanaka2023interaction,siyao2024duolando}, joint-level control~\cite{xie2024omnicontrol,karunratanakul2023gmd, TLControl}, and style editing~\cite{motion2016holden,camdm}.

Motion editing through text has evolved along two main paths: in-motion editing for specific body parts~\cite{motionllm,kim2022flame,huang2024como} and segment-level editing using text prompts. Notably, FlowMDM~\cite{flowmdm} demonstrated impressive results in seamless transitions between local motion segments. STMC~\cite{petrovich24stmc} proposed a hybrid method for spatial and temporal motion composition using pre-trained motion models. UniMotion~\cite{li2024unimotion} leveraged per-frame and sequence-level text to enhance motion understanding and control. 

While these approaches have advanced the field significantly, they typically assume simplified environments with uniform height and flat terrain. Our work extends these capabilities by incorporating complex scene geometry while maintaining text-based semantic control.

\subsection{Scene-aware Motion Synthesis.}
\begin{figure*}[ht!]
    \centering
    \includegraphics[width=.6\linewidth]{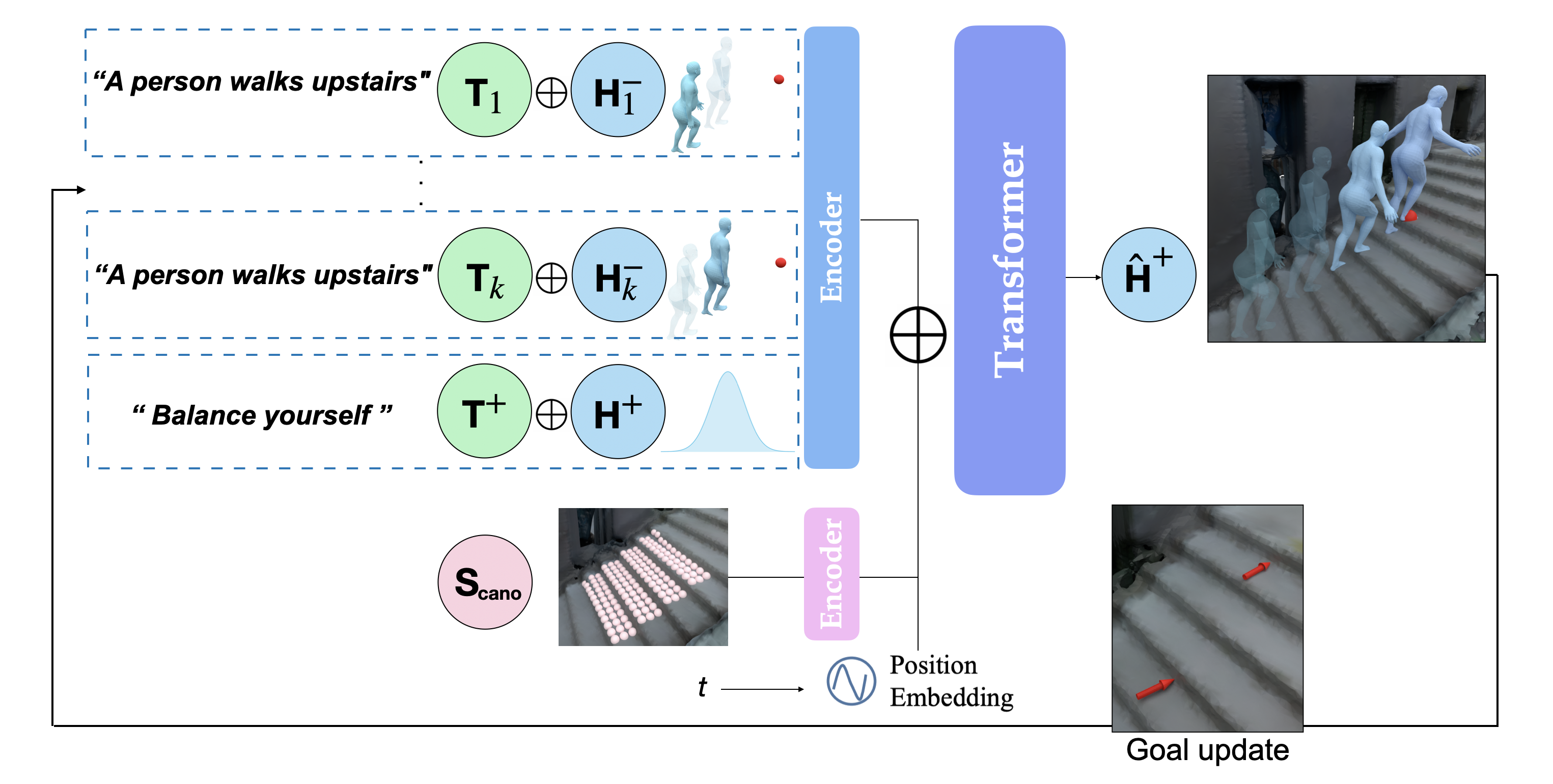}
    \caption{Architecture overview. \model has a 3D scene, a user-defined trajectory, and text prompts, and the past human motion as inputs. The past human motion and the scene encoding first undergo goal-centric canonicalization. The diffusion-based transformer then encodes the aligned text-motion tokens, scene tokens and a timestamp token to predict the canonicalized future human motion.}
    \label{fig:architecture}
\end{figure*}
Scene-aware motion synthesis is a comprehensive field that can be broadly classified into two categories: object interaction and scene navigation. Research on human-object interaction~\cite{kim2024parahome,zhang2024hoi,bhatnagar22behave,zhang2024core4d} spans a wide range, from interactions with large, static objects like chairs and beds~\cite{zhang2022couch,samp,zhang2023roam,jiang2023chairs,nsm,Pi_2023_ICCV, pan2023synthesizing,yi2024tesmo,kulkarni2023nifty}, to dynamic engagements with moving objects. This includes studies that focus on contact-based object interactions without navigation~\cite{THOR,xu2023interdiff,diller2023cghoi,xu2024interdreamer,peng2023hoi,yang2024fhoi}, as well as those that incorporate navigation~\cite{zhang2024force,li2023controllable, li2023OMOMO, li2024chois}.
A parallel line of research leverages reinforcement learning to synthesize interactions~\cite{hassan2023,merel2020catchcarry,cui2024anyskill}. Other studies have concentrated on full body grasps~\cite{taheri2021goal,Araujo_2023_CVPR,diomataris2024wandr,taheri2021goal,tendulkar2022flex, DBLP:conf/wacv/Li0L024} and dexterous hand manipulation~\cite{liu2024geneoh, christen2022dgrasp, taheri2023grip,manipnet,braun2023physically,dfbgrasp2024braun}.

In the context of human-scene interactions, a significant portion of the work is dedicated to generating short-term motion within 3D scenes~\cite{wang2022humanise,cen2024text_scene_motion,wang2024move}. PFNN~\cite{pfnn} introduced a real-time motion controller that adapts to uneven terrain but requires carefully annotated phase labels and does not enable text-based motion style editing. Some models generate longer-term human motion but often require a full-body target pose as a control signal~\cite{Zhao:ICCV:2023,liu2023mob}. Others assume uniform height within the scenes~\cite{trumans, mir23origin,lee2023lama}. Using reinforcement learning, ~\cite{luo2024universal, rempeluo2023tracepace} propose policies for terrain traversal, however, the motion is not human-like due to the animation of the physical character. Moreover, their synthesis only perform on synthetic terrains with limited complexity.

More recent work incorporates text control into human-scene interaction. TeSMO~\cite{yi2024tesmo} proposed a two-stage method for collision-free navigation within the scene. TRUMANS~\cite{trumans} unified static and dynamic object interactions, and a recent extension replaced action labels with more versatile text prompts~\cite{jiang2024autonomous}, achieving impressive results. However, these models still assume flat terrains or floors. While some concurrent works have demonstrated human motion on stairs~\cite{Zhao:DART:2024,cong2024laserhuman}, they have their limitations. Zhao et al.~\cite{Zhao:DART:2024} did not train their model with paired motion-scene data. This lack of scene awareness restricts the model's ability to generalize to complex scene constraints and adapt to changes in terrain surfaces. Moreover, their approach requires the future 3D root position, which is not always available. On the other hand, Cong et al.~\cite{cong2024laserhuman} did not enable control with the goal location, limiting its controllability and the length of plausible motion sequences it can generate.

Our work addresses these limitations by introducing the first scene-aware motion synthesis model that can adapt to the terrain and is controllable with text-based semantic signals. Our versatile model synthesizes realistic human motion across diverse 3D environments while allowing semantic control over motion style.

\section{Method}
Our proposed diffusion model generates arbitrarily long human motions that adapt to complex terrains while allowing semantic control through text prompts. The key insight is decomposing the complex task into hierarchical reasoning levels: high-level movement planning in the goal canonical frame and fine-grained scene adaptation through local geometry reasoning.

\subsection{Problem Formulation}
As illustrated in Figure~\ref{fig:architecture}, given a 3D scene, a user-defined trajectory consisting of sub-goals $\{\goal_{j}\}_{j=1}^{M}$, and text prompts $\clip$, our model is designed to fulfill both the environmental and textual constraints. It synthesizes motion $\human$ that reaches the goals, adapts to complex scene surfaces, and avoids penetration. Moreover, our motion style can be controlled by user-specified text instructions.
\subsection{Data Representations}
To synthesize scene-aware semantic motion, our method takes four key representations:
\paragraph{Human Motion $\human$} Unlike previous motion representation of human motion~\cite{tevet2023human,trumans,Guo_2022_CVPR}, which requires an additional fitting process to obtain the final animated mesh, our representation can be animated directly. The SMPL model~\cite{smpl} is used to parameterize our human motion. Our motion human $\human$ consists of $N$ frames of the joint rotations in the 6-D continuous form~\cite{zhou_6d} $\poserotation\in\mathbb{R}^{N\times22\times6}$, and the global root location $\mathbf{J}_{\textrm{root}}\in\mathbb{R}^{N\times3}$. The binary foot contact for the heel and toe joints $\contact\in\mathbb{R}^{N\times4}$ are also included.

\paragraph{Scene embedding $\scene$}
Inspired by~\cite{Chan2022eg3d}, the scene is encoded by a distance field~$
\scene\in\mathbb{R}^{N\times H\times H}$ centered at the human root joint and its orientation is relative to the Y-rotation of the root. This local representation enables efficient processing of relevant terrain features while maintaining translation invariance. The embedding is sampled by projecting from the point grid perpendicularly toward the scene. Previous approaches adopt an occupancy representation by encoding the scene with binary values~\cite{liu2023mob,cen2024text_scene_motion,trumans,nsm}. Instead, our embedding is more efficient and informative for the character to adapt to the terrain. Empirically, we use $H\times H$=144 points that are uniformly sampled from a $1.2\times1.2$ meter grid. 

\paragraph{Goal Representation}
Each sub-goal $\goal_{j}$ to our system is represented by a target 3D position to be reached on the scene $\goalpos^{j}\in\mathbb{R}^{3}$, and a 2D desired orientation vector represented by $\goaldir^{j}\in\mathbb{R}^{2}$. 

\paragraph{Text Control $\clip$} Unlike previous methods that use a single text embedding combined with a timestamp~\cite{shafir2023human, tevet2023human, shafir2024human}, we employ a different approach. We encode the text on a per-frame basis and treat each frame's text as an individual token within the diffusion transformer. This method of temporal tokenization ensures a precise alignment between the motion and the corresponding text~\cite{li2024unimotion}, facilitating a seamless transition between different motion styles. The text prompt $\clip\in\mathbb{R}^{N\times D}$ is obtained by reducing the dimensionality of the CLIP embeddings using PCA. In our experiments, the CLIP embedding is reduced to $D=64$ dimensions.

\subsection{Goal-Centric Canonicalization}
\label{subsec:cano}
One key to our model is the goal-centric canonicalization that ensures robust goal-reaching, while maintaining physical plausibility. This transformation serves two crucial purposes: (1) it simplifies the learning problem by creating a consistent reference frame for motion synthesis, and (2) it enables better generalization across different goal configurations. We transform both our human motion and scene embedding ($\human$ and $\scene$) into the coordinate system of the goal so that the model can combine the high-level reasoning of the goal and the fine-grained reasoning of the complex scene geometry. First, under current goal $\goal_j$, we apply canonicalization to the motion $\human$ via
$\human_{\textrm{cano}} = \mathcal{T_\textrm{human}}(\human, \goal_j).$
 Traditional methods~\cite{trumans,jiang2024autonomous,yi2024tesmo}, which explicitly condition on the goal, can often lead to inaccuracies in reaching the target. Our experiments show that this is accentuated when synthesizing motion on uneven terrain surfaces. Therefore, the model is instead trained to synthesize motion that converges to the origin in the coordinate system defined by the goal. Moreover, the scene embedding is transformed to align with the height of the goal via $\scene_{\textrm{cano}} = \mathcal{T_\textrm{scene}}(\scene, \goal_j)$. This way, the model does not only implicitly learn to reason about the goal, but also becomes aware of the local scene geometry. Additionally, the current height of the root is encoded.

\subsection{Autoregressive Motion Diffusion}
The synthesis process seamlessly connects multiple motion segments through an autoregression. As shown in Figure~\ref{fig:architecture}, each segment is predicted using the previous one, maintaining continuity while adapting to new goals and terrain features. The model synthesizes scene-aware motion towards the current sub-goal $\goal_j$. Once the sub-goal is reached, the goal iterates to $\goal_{j+1}$. This way, the model can progressively synthesize arbitrarily long motions that are plausible to the scene. Such an approach not only enables the length of the animation to become unconstrained, but also allows users to control the motion trajectory to avoid obstacles. 

\paragraph{Conditional Diffusion Model}
Each motion segment is generated through a conditional diffusion process, which incorporates a transformer architecture, as depicted in Figure~\ref{fig:architecture}. The generation of successive segments is facilitated by using the last $k$ frames of the preceding segment as a seed motion, which then extends to the next segment. We denote the canonicalized motion segment $\human_{\textrm{cano}}$ defined in Sec~\ref{subsec:cano} as a combination of the $k$ frames of seed motion $\human^{-}$, and the $N-k$ frames of predicted motion $\human^{+}$. The diffusion process is conditioned on several factors: the scene embeddings $\scene$, the text prompt $\clip$, and the past seed motion, $\human^{-}$. Together, these are represented as the condition, $\condition=(\scene, \clip, \human^{-})$. In our experiments, we set the values of $N$ and $k$ to 40 and 10, respectively. During the training phase, noise is injected into the future motion, $\human^{+}$, while the seed motion, $\human^{-}$, remains unchanged. At each denoising step $n$, the model learns to reverse the forward diffusion process, with the reverse process defined as
\begin{equation}
    p(\human^{+}_{n-1}|\human^{+}_n, \condition) := \mathcal{N}(\human^{+}_{n - 1}; \mu(\human^{+}_{n} ,\condition), \Sigma_n),
\end{equation}
where $\mu$ denotes the predicted mean and $\Sigma_n$ is a fixed variance. Learning the mean can be re-parameterized as learning to predict the clean future motion $\human^{+}_{0}$. During training, we also apply an $l_2$ loss on the predicted joint positions obtained via forward kinematics:
\begin{equation}
    \mathcal{L} = \mathbb{E}_{\human^{+}_0}\|\hat{\human}^{+}_0 - \human^{+}_0\|_2 + \lambda\cdot\|\hat{\poseposition}^{+} - \poseposition^+ \|_2.
\end{equation}
This is crucial for the sharpness of the motion. Here, $\hat{\human}^{+}_0$ denotes the predicted future motion, while $\hat{\poseposition}^+$ denotes the predicted future joint positions obtained via forward kinematics. The positional loss weight $\lambda$ is set to be 4.

\subsection{Object Interaction}
When the human arrives in the vicinity of the target object after the navigation, our method generates full-body motion by interacting with the objects to perform text-controlled sitting and lying. 
Instead of focusing on the goal and the neighboring scene, the interaction model needs to be aware of the target object geometry. For this reason, we introduce another diffusion model conditioned on an object geometric representation~$\object\in\mathbb{R}^{2048}$. The representation comprises the distances from the basis point set (BPS)~\cite{bps} to the object surface, as well as the distance from the hands and the hip joints to each one of the object voxels. The BPS consists of $512$ points uniformly sampled from a sphere of radius $1$~meter, centered around the normalized object center. The object is voxelized into an $8\times 8\times 8$ grid, and we zero out the distance features for unoccupied voxels. The interaction model employs the same representation for human motion and texts. We train our interaction model on the SAMP~\cite{samp} dataset. The interaction diffusion model is trained using the same learning objective as the navigation model.

\subsection{Scene-aware Guidance}
At test time, diffusion models can be guided to meet specific objectives, alleviating the need for training models with different configurations, and further enhancing the quality of scene interaction. For the sake of readability, we will use the same notation for both estimated and true values in the following discussion. We directly apply the guidance to the clean motion prediction from the model $\human_{0}$~\cite{ho2022video,yi2024tesmo,li2024chois,karunratanakul2023gmd}. At each denoising step, the predicted $\human_{0}$~is updated with the gradient of an analytic objective function $\mathcal{J}$. This process can be denoted as $\tilde{\human}_{0} = \human_{0} - \alpha\Delta_{\human_{t}} \mathcal{J}(\human_{0})$, where $\alpha$ controls the strength of the guidance and $\human_{t}$ is the noisy input motion at diffusion step $t$. The predicted mean $\mu$ is then calculated
with the updated motion prediction $\hat{\human}_0$ 

For navigation, we further introduce a physics plausibility guidance to avoid penetration and encourage realistic contact. By enforcing foot contact when it happens and penalizing the foot penetration with the scene when there is no contact. Formally, the guidance is computed by
\begin{equation}
    \mathcal{J}_{\text{phys}} = \contact \cdot \|\mathbf{J}_{\text{feet}} - \mathbf{h} \|_2  + (1 - \contact) \cdot \mathbbm{1}(\mathbf{h}  > \mathbf{J}_{\text{feet}}) \cdot \|\mathbf{J}_{\text{feet}} - \mathbf{h} \|_2.
\label{eqn:err}
\end{equation}
Here, we leverage the predicted foot contact label $\contact$ to enforce accurate foot contact with the scene and to discourage penetration. Furthermore, we denote the predicted foot joint positions as $\mathbf{J}_{feet}$ and the heights of the projected points from the feet as $\mathbf{h}$.

For the interaction model, a collision objective is used to discourage penetrations~\cite{li2024chois,yi2024tesmo} between humans and objects $\mathcal{J}_{\textrm{collision}} = \textrm{SDF}(v)$, where object signed distance field (SDF) is queried by the body vertices $v$, and the mean penetration distance of the body vertices is minimized. In addition to the object collision guidance, we also incorporate a motion smoothness objective $\mathcal{J}_{\textrm{smooth}} = \|\mathbf{J}_p^{1:N} - \mathbf{J}_p^{0:N-1} \|_2$

For the navigation model, we set the guidance weight $\alpha$ to 3 for physics guidance and 50 for smoothness guidance. For the interaction
model, we utilize weights of 50 for the collision guidance. To ensure smooth generation results, we apply the inference guidance at the final time step of denoising. For a fair comparison with baselines, the inference guidance is not activated for all comparisons.

\section{Experiments}

\begin{table*}[ht]
    \centering
    \caption{Quantitative evaluations against baseline methods, and ablation study on key components and design.}
    \renewcommand{\arraystretch}{1.05}  
    \setlength{\tabcolsep}{4pt}  
    \resizebox{\textwidth}{!}{%
    	\begin{tabular}{l|cccccccccc}
            \cline{1-11}
            \multirow{2}{*}[-3pt]{\textbf{Methods}} & \multicolumn{2}{c}{\hspace{1em}\textbf{Scene constraints}} & \multicolumn{2}{c}{\hspace{.5em}\textbf{Goal reaching}} & \multicolumn{4}{c}{\textbf{Motion quality}} & \multicolumn{1}{c}{\hspace{1em}\textbf{User Study}($\%$)} \\
            \cmidrule(lr){2-3}\cmidrule(lr){4-5}\cmidrule(l){6-9}
            & Penetration↓ & Contact Dist.↓ & Pos.↓ & Rot.↓ & FID↓ & Multimodality→ & Diversity→ & Foot-skate↓ & \\
            \cline{1-11}
            Ground Truth & - & - & - & - & 0.000 & 6.023 & 12.410 & - & -\\
            FlowMDM*~\cite{flowmdm} & 4.67 & 6.94 & 4.79 & 0.125 & 66.485 & 9.107 & 17.038 & 2.949 & 9.5 \\
            TRUMANS*~\cite{trumans} & 4.50 & 6.65 & 3.38 & 0.0454 & 26.533 & 8.172 & 14.717 & 3.329 & 14.9  \\
            \cline{1-11}
            Ours no cano. & 1.98 & 5.55 & 3.51 & 0.0796 & 8.021 & 7.344 & 13.507 & 2.710 & -\\

            Ours no scene emb. & 2.99 & 5.74 & 1.57 & 0.0384 & 1.924 & \textbf{5.823} & \textbf{12.519} & 2.678  & - \\
            \cline{1-11}
            Ours & \textbf{1.57} & \textbf{4.51} & \textbf{1.38} & \textbf{0.0376} & \textbf{1.680} & 6.354 & 13.067 & \textbf{2.671}  & \textbf{75.6}\\
            \cline{1-11}
        \end{tabular}
    }
    \label{tab:my_label}
\end{table*}

\begin{figure*}[h!]
    \centering
    \includegraphics[width=.85\textwidth]{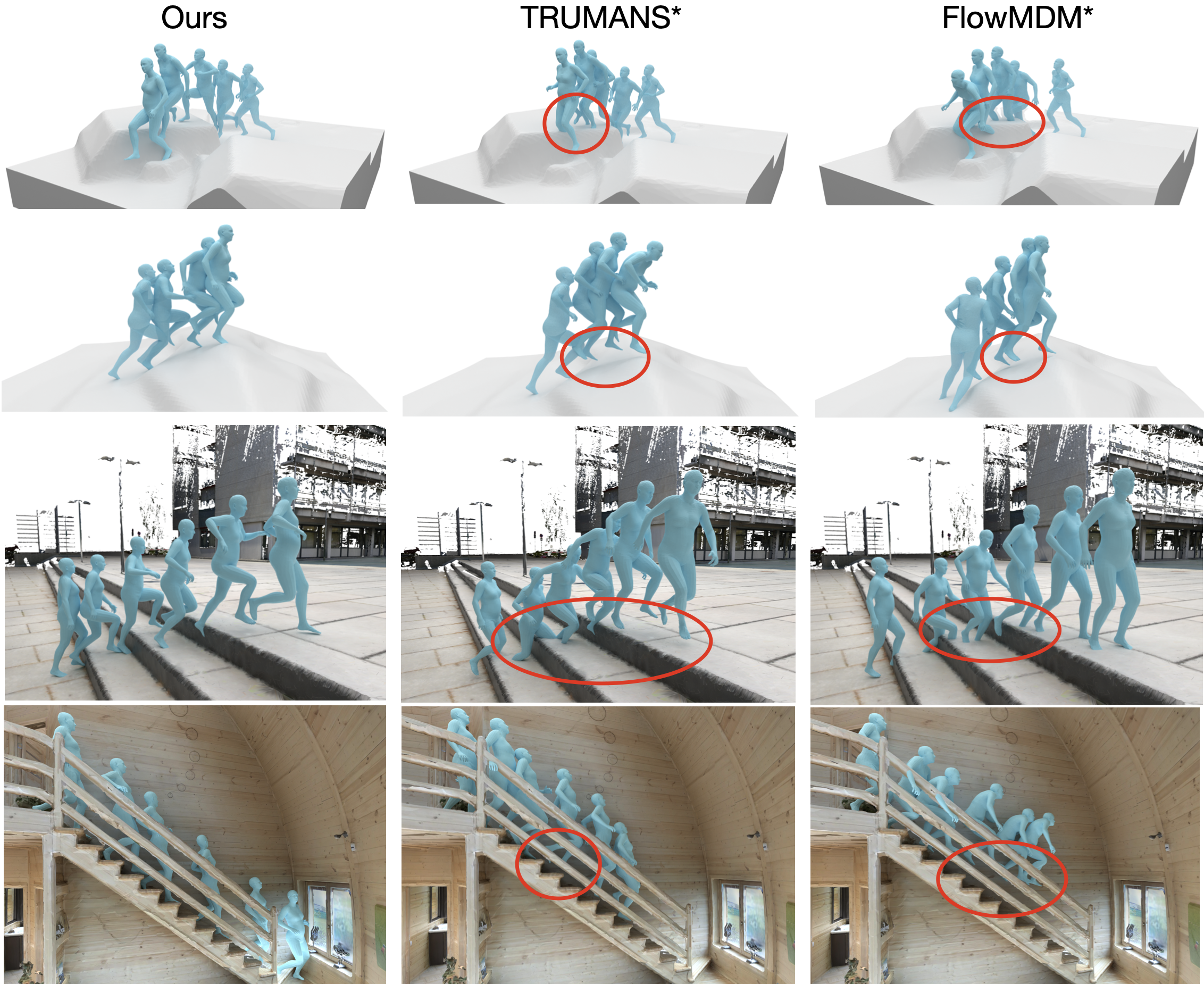}
    \caption{Qualitative comparison with baselines. Results are on the testing set of the~\model dataset (top two rows). Without the hierarchical reasoning of the scene, the baseline methods produce more penetration with the legs (first row) and the floating effect (second row). Furthermore, our method generalizes to real-world scene datasets of HPS~\cite{HPS} and MatterPort3D~\cite{Matterport3D} (bottom two rows)}
    \label{fig:qualitative}
    \vspace{-8pt}
\end{figure*}
\begin{figure*}[ht!]
    \centering
    \includegraphics[width=0.65\textwidth]{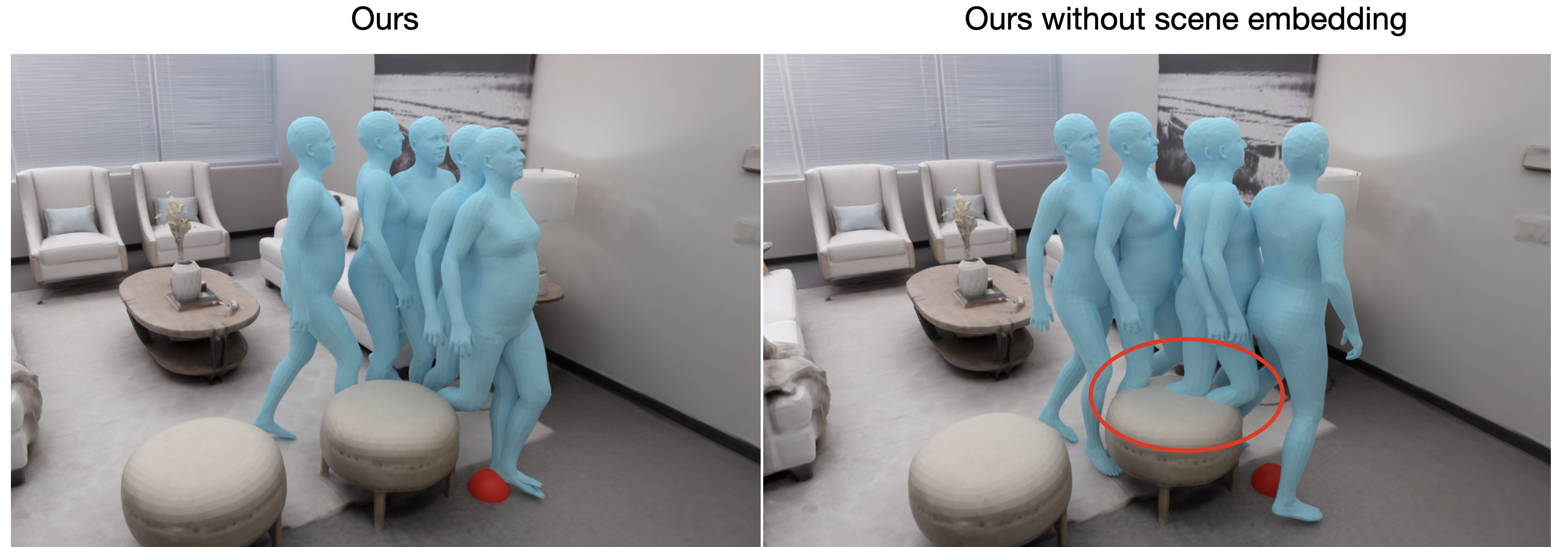}
    \caption{Ablation on the human-centric scene embedding. It is significant in preventing unwanted interactions with cluttered environments.}
    \label{fig:ablation}
    \vspace{-8pt}
\end{figure*}

\begin{figure*}[ht!]
    \centering
    \includegraphics[width=\textwidth]{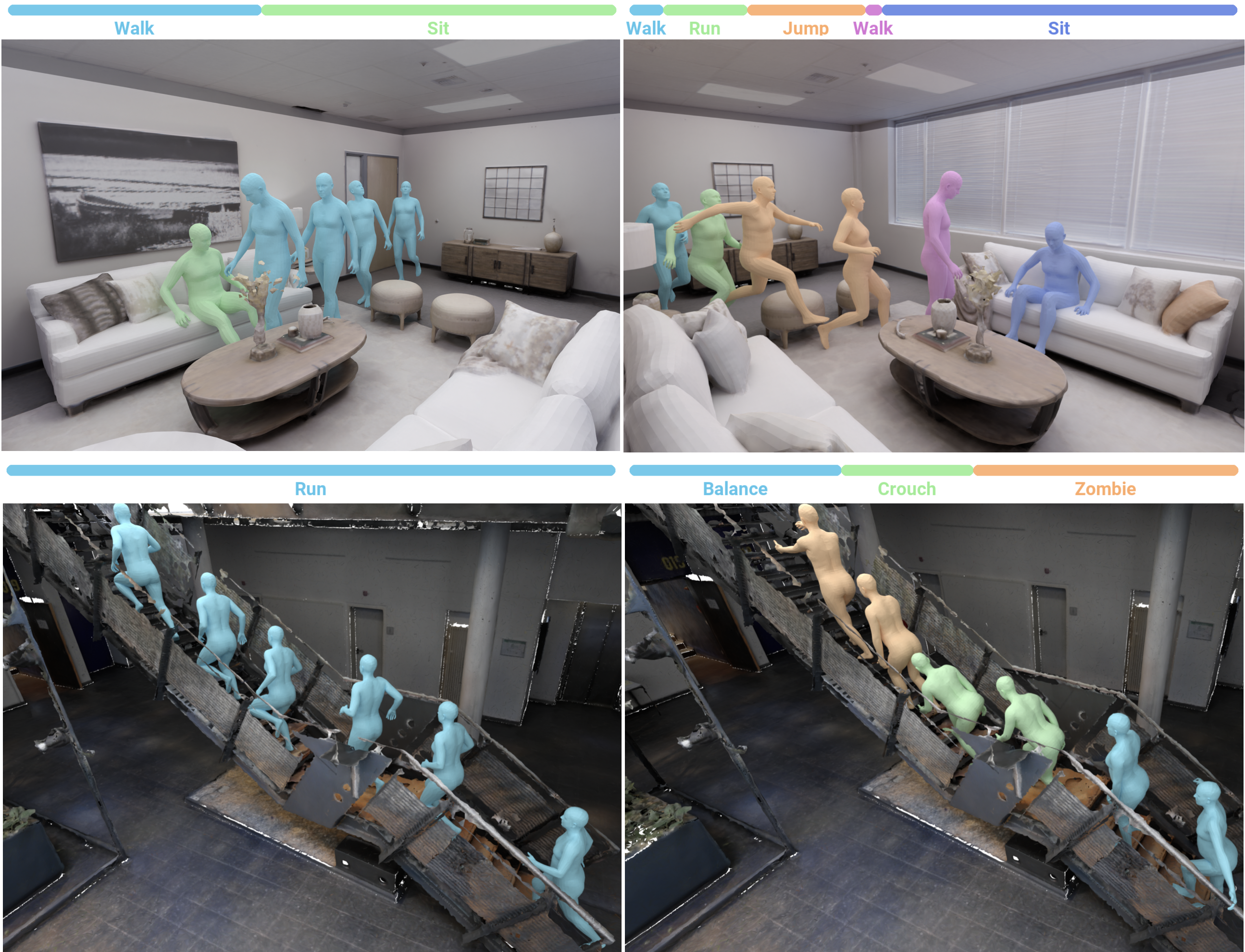}
    \caption{\model generalizes to novel scenes and text instructions, as demonstrated with Replica~\cite{replica19arxiv} and HPS~\cite{HPS} scenarios. The model follows instructions like \textit{take a walk}, \textit{sit on the sofa}, and \textit{run up the stairs}, and adapts to more complex commands such as \textit{jump over a stool} while adjusting to scene constraints. In the HPS scene, the model transits between different gait styles, following the text control while adapting to the staircases.}
    \label{fig:edit}
    \vspace{-8pt}
\end{figure*}

First we introduce our dataset and evaluation metrics. Then we show comparisons of our proposed approach against the baselines. We further conduct a human perceptual study to complement our evaluation and ablation study to verify the effectiveness of our key components.

\subsection{Dataset and Implementation Details}
\paragraph{The~\model Dataset} To our knowledge,~\cite{cong2024laserhuman,jiang2024autonomous} are the only existing dataset that captures human navigation with scenes and text annotations. However, both its motion style and terrain variation are limited. \\To address the scarcity of paired human-scene-text data, we utilize a vast database of artificial heightmaps~\cite{pfnn}, derived from video game environments. This approach allows us to match human motion segments with the most suitable terrain patches, thereby generating paired human and scene data. We divide the motion sequences into clips of 60 frames (2 seconds) each, aligning the human's initial position with the center of the $4\times4$ meter patches. The terrains with minimized foot contact and penetration error are retrieved, where the error is computed similarly to Equation~\ref{eqn:err}. To diversify our dataset, we record motion featuring various motion styles across different terrains. Our motion set includes a dataset captured with Inertial Motion Units (IMUs) and the PFNN~\cite{pfnn} motion dataset retargeted to the SMPL format. 
The dataset comprises 15000 sequences, and 1000 sequences are reserved for testing. To augment our data, pose mirroring is performed along the x-axis and for each motion sequence. Three best-fitted terrains are used for training. 

\paragraph{Implementation Details}
All models including the baselines are trained for 400k steps. The navigation models are trained on the~\model dataset and the interaction model is trained on our text-annotated SAMP~\cite{samp}. All models are trained to denoise the input in 100 diffusion steps.

\subsection{Baselines}
We train all the baselines and perform an ablation study on the~\model dataset. We compare our work with state-of-the-art diffusion-based methods. TRUMANS~\cite{trumans} achieves impressive performance for scene interaction, since it does not condition on text prompts, we replace its action encoding with a text encoding. This text-variant of TRUMANS is denoted as TRUMANS*. FlowMDM~\cite{flowmdm} does not consider the surrounding scene, we enhance its scene awareness by additionally incorporating the same occupancy representation that was adopted in the original TRUMANS model. \\
To justify our key hierarchical scene reasoning, ablation is performed on the goal-centric canonicalization, where instead the motion is canonicalized to the first frame and the goal is provided explicitly. Another baseline is introduced to evaluate the importance of the local scene reasoning by not incorporating the scene embedding.

\subsection{Evaluation Metrics}
An important aspect of assessing the model is to evaluate how well it satisfies the scene constraint. \textbf{Penetration} (cm) measures the average penetration distance for all the human body vertices~\cite{yi2024tesmo,yi2022mime,li2024chois,trumans}, obtained by querying all body vertices from the computed SDF of the testing scenes. \textbf{Contact distance} (cm) evaluates the average distance to the scene when there is contact. For this, we annotate four body vertices - one at the toe and the heel of each foot. \\
For goal reaching, we evaluate the body-to-goal \textbf{positional} (cm) and \textbf{rotational offset} (radians).~\cite{yi2024tesmo,Zhao:DART:2024}. \\ We follow~\cite{guo2020action2motion,tevet2023human,yi2024tesmo,trumans,li2024chois,Zhao:DART:2024} and evaluate the motion embeddings of an action recognition model~\cite{stgcn2018aaai,cong2024laserhuman} trained on the~\model dataset with all ten action classes. \textbf{Multimodality} measures the alignment between the generated motion and the text instruction. \textbf{Frechet Inception Distance} (FID)~\cite{guo2020action2motion} measures the realism of the motion compared to the ground truth. \textbf{Diversity} is computed based on the average pairwise distance between sampled motions.
\vspace{-1em}
\paragraph{Human Perceptual Study}
In addition to the quantitative measures introduced, we also conducted a user study on the realism as well as the controllability of the methods through text. In the user study, we presented animations on the real-world scenes from HPS~\cite{HPS} and Matterport~\cite{Matterport3D} to 24 participants. The participants make three-way comparisons of animations generated by the three methods in shuffled order. We have incomplete responses filtered out. Details of the user study can be found in the supplementary. 

\subsection{Quantitative Evaluation.}
From Table~\ref{tab:my_label}, our model achieves competitive performance across all evaluation metrics compared to baseline methods. In terms of scene constraints, our approach attains the lowest penetration (1.57 cm) and contact distance (4.51 cm), outperforming FlowMDM* and TRUMANS*. In goal-reaching, our method exhibits the best performance in both positional accuracy (1.38 cm) and rotational alignment (0.0376 radians). This validates our design choice of goal-centric canonicalization. Regarding motion realism, our approach achieves the lowest FID score (1.680) among all compared methods, being closest to the ground truth. Our method maintains comparable diversity (13.067) and multimodality (6.354) scores close to the ground truth distribution (12.410 and 6.023 respectively). Our model also produces the least foot-skate artifact (2.671 cm). In the user study, 75.6\% of participants preferred~\model over the baselines. This strong preference confirms our method's effectiveness in generating visually plausible human-scene interactions, particularly in reducing floating and penetration artifacts, while generating realistic contacts.


\subsection{Qualitative Evaluation}
We present qualitative comparisons in Figure~\ref{fig:qualitative}. The top two rows demonstrate results from the~\model dataset's test set, where baseline methods exhibit noticeable artifacts - leg penetration into the ground surface - due to their limited scene understanding. In contrast, our approach, leveraging hierarchical scene reasoning with scene embedding and goal-centric canonicalization, generates motions that maintain proper contact while avoiding both penetration and floating artifacts. The bottom two rows highlight the generalization capabilities of our approach across different scene datasets, namely MatterPort3D~\cite{Matterport3D} and HPS~\cite{HPS}. These real-world environments pose more diverse and challenging scenes than those in our training set. Despite these complexities, our method consistently generates physically plausible motions that adhere to scene constraints across these varied terrains. This robust performance again stems from our hierarchical scene reasoning. These results demonstrate that our method not only excels in controlled test scenarios but also effectively adapts to novel, real-world environments. Please refer to our supplementary video for results and comparisons in motion.

\subsection{Ablation}
The usefulness of our core components of goal-centric canonicalization and human-centric scene embedding are justified through the comparison with the ablative baselines. For goal-reaching capability, it is highlighted in Table~\ref{tab:my_label}, where our method (1.38 cm, 0.0376 radians) achieves better performance over the baseline without canonicalization (3.51 cm, 0.0796 radians) validates our design choice of goal-centric canonicalization.

In regards to scene awareness, it is illustrated in Figure~\ref{fig:ablation} that without the scene embedding, the model is more likely to exhibit unwanted penetrations with the cluttered scenes while navigating. With the scene embedding, our model avoids the tea table in the way of reaching the sub-goal. It can navigate while following the sub-goal. The importance of scene awareness is further supported by Table~\ref{tab:my_label}, shown in the improvement over our ablation without the scene embedding (2.99 cm penetration) particularly emphasizing the importance of local scene reasoning in preventing body-scene intersections.

\subsection{Generalization}
\model is capable of generalizing to both novel real-world scenes and text instructions. As shown in Figure~\ref{fig:edit},~\model navigates in Replica~\cite{replica19arxiv} and HPS~\cite{HPS} 
The model is firstly instructed to ``take a walk'' before ``sitting on the sofa" (top left) and ``running up the stairs'' (bottom left). In more complicated scenarios, the model adapts to the scene constraints while following the ``jump over a stool" instruction, before ``sitting on the sofa" (top right). In the HPS scene, the human transits between various gait styles controlled by text while adapting to the stairs. Similarly in Figure~\ref{fig:teaser},~\model is provided a series of text instructions before lying on the sofa in the LaserHuman scene~\cite{cong2024laserhuman}.

\section{Conclusion}
We presented~\model\removespace, the first diffusion-based motion synthesis model that simultaneously enables text-controlled style editing and adaptation to complex terrains. Our model introduces several key technical innovations, including a goal-centric canonical coordinate frame for long-term navigation and a hierarchical scene reasoning approach that combines high-level goal understanding with fine-grained scene awareness. Through extensive experiments across multiple scene datasets, we demonstrated that our approach significantly outperforms existing methods, achieving the best performance in both scene constraint satisfaction and motion quality. User studies further validate our approach, having 75.6\% of the participants preferring our method over state-of-the-art methods.\\
In the future, this work can be extended to more complex scene interactions. Directions include incorporating dynamic object manipulation during navigation, such as carrying objects while climbing stairs. Additionally, incorporating collision avoidance mechanisms for dynamic and cluttered environments would benefit real-world applications of virtual social interaction and autonomous driving.

\section*{Acknowledgement}
A big thank you goes to Hongwei Yi for the useful discussions and exchanging ideas.  We appreciate the RVH group members for their useful feedback. This work is funded by the Deutsche Forschungsgemeinschaft (DFG, German Research Foundation) - 409792180 (Emmy Noether Programme, project: Real Virtual Humans), and German Federal Ministry of Education and Research (BMBF): T{\"u}bingen AI Center, FKZ: 01IS18039A, and Huawei Noah's Ark Lab. Gerard Pons-Moll is a member of the Machine Learning Cluster of Excellence, EXC number 2064/1 – Project number 390727645. The project was made possible by funding from the Carl Zeiss Foundation.
\newpage

\bibliographystyle{splncs04}
\bibliography{Main}
\newpage

\section*{APPENDIX}

\begin{figure*}[ht!]
    \centering
    \includegraphics[width=0.75\textwidth]{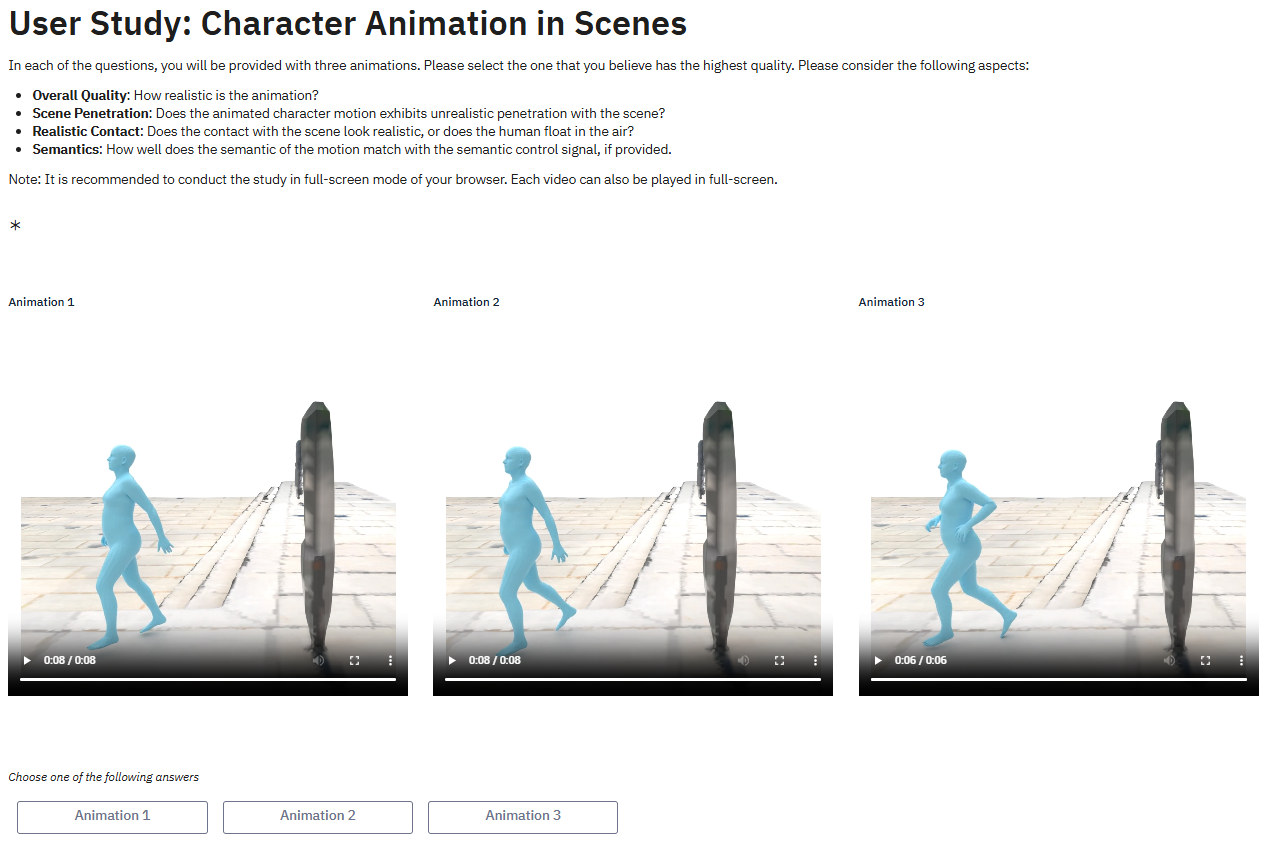}
    \caption{The layout of our perceptual study for evaluating perceived realism, compliance of scene constraints, and text-based controllability of~\model.}
    \label{fig:ablation}
    \vspace{-8pt}
\end{figure*}

\section*{1. Details on User Study}
Our evaluation encompasses a human perceptual study, which is aimed at assessing both the ability of our methods to satisfy scene constraints and their controllability through text. We utilized animations derived from the HPS~\cite{HPS} and Matterport~\cite{Matterport3D} datasets for this purpose. Each participant was presented with a set of seven questions, as illustrated in Figure~\ref{fig:teaser}, requiring them to perform a three-way comparison of animations. These animations were presented in a randomized order to prevent any ordering bias.

The study received 24 complete responses for the final analysis. The results were encouraging, with 75.6\% of participants expressing a preference for our model over the baseline alternatives. This strong preference highlights the effectiveness of our method in generating believable human-scene interactions. Notably, our approach significantly reduces floating and penetration artifacts while promoting the generation of realistic contacts.

Overall, our user study validates the effectiveness of our method in creating visually plausible animations that adhere to scene constraints and can be manipulated through text.

\section*{2. Dataset}
\label{sec:data}
\subsection*{2.1. Terrain Fitting Process}
Since capturing simultaneously human motion with scenes that include diverse terrains is expensive and difficult, we leverage a method that fits 2 second motion segments (60 frames) onto a set of 20,000 4x4 meters terrain patches to obtain paired motion-scene data. The terrain patches are sampled at random locations and orientations from large terrain scenes from Source Engine. By leveraging ray-tracing, the full geometric information are encapsulated in the form of heightmaps with a resolution of one pixel per inch. We then construct the patched terrain heightmaps into watertight meshes.

Having sampled the terrain meshes, the motion segments are then fitted in two main stages:
\begin{enumerate}
    \item \textbf{Patch Selection:} Identify the three best-matching terrain patches using a brute-force search that minimizes a comprehensive error function.
    
    \item \textbf{Terrain Refinement:} Apply a Radial Basis Function (RBF) mesh editing technique to ensure precise foot placement accuracy.
\end{enumerate}
The error function $E_\text{fit}$ comprises three key components: $E_\text{contact}$ ensures foot height matches ground contact point. $E_\text{penetration}$ prevents intersection when feet are not in contact with the terrain. $E_\text{jump}$: is only activated when the character is jumping, ensuring the height of the terrain is no more than $l$ in distance below the feet.

\begin{equation}
E_{\text{fit}} = E_{\text{contact}} + E_{\text{penetration}} + E_{\text{jump}}
\end{equation}

\begin{equation}
E_{\text{contact}} = \sum_i \sum_{j \in J} \contact_j^i (\mathbf{h}_j^i - \mathbf{J}_{\text{feet},j}^i)^2
\end{equation}

\begin{equation}
E_{\text{penetration}} = \sum_i \sum_{j \in J} (1 - \contact_j^i) \max(\mathbf{h}_j^i - \mathbf{J}_{\text{feet},j}^i, 0)
\end{equation}

\begin{equation}
E_{\text{jump}} = \sum_i \sum_{j \in J} \mathbbm{1}_{\text{jump}}^i (1 - \contact_j^i) \max((\mathbf{J}_{\text{feet},j}^i - l) - \mathbf{h}_j^i, 0)
\end{equation}
Here,
\begin{itemize}
    \item $J_{\text{foot}}$: Set of joint indices (left/right heel and toe)
    \item $c_j^i$: Contact label for foot joint $j$ at frame $i$
    \item $f_j^i$: Foot joint height at frame $i$
    \item $h_j^i$: Terrain height under foot joint at frame $i$
    \item $\mathbbm{1}_{\text{jump}}^i$: Binary indicator for jumping gait
    \item $l$: Height threshold (approximately 0.3m)
\end{itemize}

After computing the fitting error for all terrain patches, we select the 3 patches with the lowest error for further processing. The motion are already well-fitted to the terrains. The further refinement stage involves editing the heightmap to ensure precise foot contact with the ground during contact phases. We use a simplified terrain deformation technique based on Botsch and Kobbelt et al.~\cite{DBLP:journals/cgf/BotschK05}, applying a 2D Radial Basis Function (RBF) with a linear kernel to the terrain fit residuals. This approach provides a flexible method for adapting character motion to varied terrain geometries, multiplying the effectiveness of data and enables training generalizable models.

\subsection*{2.2. Dataset Statistics}

Our dataset includes ten gait motion styles with annotated text prompts and corresponding terrain scene patches. Table~\ref{table:data} details the dataset's motion style distribution, encompassing various locomotion types from walking and running to more specialized movements like climbing and balancing.

\begin{table}[ht!]
    \small
    	\caption{Detailed statistics of the~\model dataset. The dataset comprises 3 hours of motion (at 30fps), texts annotations, and fitted terrain meshes.} 
	\centering
    \begin{tabular}[b]{c|cc}
       \hline
       {Gait} & {Minutes} & {$\%$}  \\
       \hline
       $\textrm{Stand}$ & 6.88 & 4.09 \\
       $\textrm{Walk}$ & 75.95 & 45.17  \\
       $\textrm{Run}$ & 50.75 & 30.18 \\
       $\textrm{Crouch}$ & 14.06 & 8.36  \\
       $\textrm{Climb}$ & 2.30 & 1.37  \\        
       $\textrm{Jump}$ & 10.01 & 5.95  \\
       $\textrm{Hop}$ & 2.54 & 1.51  \\ 
       $\textrm{Balance}$ & 2.69 & 1.60  \\
       $\textrm{Zombie}$ & 2.91 & 1.73  \\
       $\textrm{Push}$ & 0.07 & 0.04  \\
       \hline
    \end{tabular}
    \label{table:data}
\end{table}

\end{document}